# Most Relevant Explanation: Properties, Algorithms, and Evaluations

**Changhe Yuan & Xiaolu Liu**
Uncertainty Reasoning Laboratory
Dept of Computer Science and Engineering
Mississippi State University
Mississippi State, MS 39762

**Tsai-Ching Lu**
HRL Laboratories, LLC
Malibu, CA, 90265

**Heejin Lim**
Uncertainty Reasoning Laboratory
Dept of Computer Science and Engineering
Mississippi State University
Mississippi State, MS 39762

## Abstract

Most Relevant Explanation (MRE) is a method for finding multivariate explanations for given evidence in Bayesian networks [12]. This paper studies the theoretical properties of MRE and develops an algorithm for finding multiple top MRE solutions. Our study shows that MRE relies on an implicit soft relevance measure in automatically identifying the most relevant target variables and pruning less relevant variables from an explanation. The soft measure also enables MRE to capture the intuitive phenomenon of explaining away encoded in Bayesian networks. Furthermore, our study shows that the solution space of MRE has a special lattice structure which yields interesting dominance relations among the solutions. A K-MRE algorithm based on these dominance relations is developed for generating a set of top solutions that are more representative. Our empirical results show that MRE methods are promising approaches for explanation in Bayesian networks.

## 1 Introduction

Bayesian networks offer compact and intuitive graphical representations of uncertain relations among the random variables of a domain and provide a foundation for many diagnostic expert systems. However, these systems typically focus on disambiguating single-fault diagnostic hypotheses because it is hard to generate "just right" multiple-fault hypotheses that contain only the most relevant faults. Maximum a Posteriori (MAP) assignment and Most Probable Explanation (MPE) are two explanation methods for Bayesian networks that find a complete assignment to a set of target variables as the best explanation for given evidence and can be applied to generate multiple-fault hypotheses. A priori, the set of target variables is often large and can be in tens or even hundreds for a real-world diagnostic system. Given that so many variables are involved, even the best solution by MAP or MPE may have an extremely low probability, say in the order of $10^{-6}$. It is hard to make any decision based on such hypotheses.

In real-world problems, it is observed that usually only a few target variables are most relevant in explaining any given evidence. For example, there are many possible diseases in a medical domain, but a patient can have at most a few diseases at one time, as long as he or she does not delay treatments for too long. It is desirable to find diagnostic hypotheses containing only those relevant diseases. Other diseases should be excluded from further tests or treatments. In a recent work, Yuan and Lu [12] propose an approach called *Most Relevant Explanation* (MRE) to generate explanations containing only the most relevant target variables for given evidence in Bayesian networks. Its main idea is to traverse a trans-dimensional space containing all the partial instantiations of the target variables and find one instantiation that maximizes a relevance measure called *generalized Bayes factor* [3]. The approach was shown in [12] to be able to find precise and concise explanations. This paper provides a study of the theoretical properties of MRE and offers further evidence for its validity. The study shows that MRE relies on an implicit soft relevance measure that enables the automatic identification of the most relevant target variables and pruning of less relevant variables from an explanation. Furthermore, the solution space of MRE has a special lattice structure that allows two interesting dominance relations among the solutions to be defined. These dominance relations are used to design and develop a K-MRE algorithm for finding a set of top explanations that are more representative. Our empirical results show that MRE methods are promising approaches for explanation in Bayesian networks.

The remainder of the paper is structured as follows. We first review methods for explanation in Bayesian networks, including Most Relevant Explanation. Then we introduce several theoretical properties of Most Relevant Explanation. We also develop a K-MRE algorithm for generating multiple top explanations and evaluate it empirically.

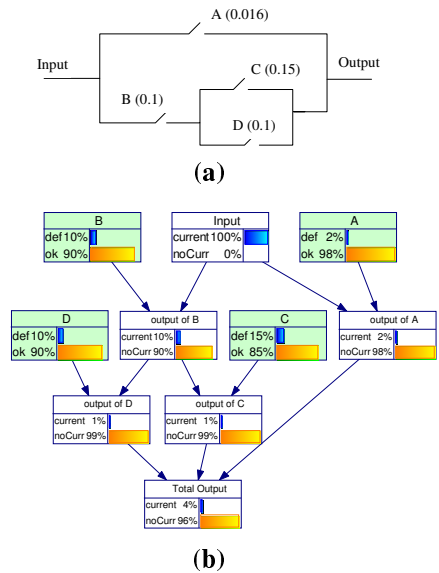

Figure 1: **(a)** A probabilistic digital circuit and **(b)** a corresponding diagnostic Bayesian network

## 2 A Running Example

Let us first introduce a running example used throughout this paper. Consider the circuit in Figure 1(a) adapted from [9, 12]. Gates $A, B, C$ and $D$ are defective if they are closed. The prior probabilities that the gates close independently are $0.016, 0.1, 0.15$ and $0.1$ respectively. Connections between the gates may not work properly with certain small probabilities. The circuit can be modeled with a diagnostic Bayesian network as shown in Figure 1(b). Nodes $A, B, C$ and $D$ correspond to the gates in the circuit and each has two states: "defective" and "ok". Others are input or output nodes and have two states: "current" or "noCurr". Uncertainty is introduced to the model such that an output node is in state "current" with a certain probability less than $1.0$ if its parent gate, when exists, is "defective" and any of its other parents is in state "current". Otherwise, it is in "noCurr" state with probability $1.0$. For example, node *output of B* takes state "current" with probability $0.99$ if parent gate $B$ is in state "defective" and parent *Input* is in state "current".

Suppose we observe that current flows through the circuit, which means that nodes *Input* and *Total Output* in the Bayesian network are both in the state "current". The task is to diagnose the system and find the best fault hypotheses. Based on our knowledge of the domain, we know there are three basic scenarios that most likely lead to the observation: (1) $A$ is defective; (2) $B$ and $C$ are defective; and (3) $B$ and $D$ are defective.

## 3 Related Work

Many methods exist for explaining evidence in Bayesian networks. However, they often fail to find "just-right" explanations containing the most relevant target variables.

Many existing methods make simplifying assumptions and focus on singleton explanations [5, 7]. However, singleton explanations may be *underspecified* and are unable to fully explain given evidence. For the running example, the posterior probabilities of $A, B, C$, and $D$ failing independently are $0.391, 0.649, 0.446$, and $0.301$ respectively. Therefore, $(\neg B)$ is the best singleton explanation[1]. However, $B$ alone does not fully explain the evidence. $C$ or $D$ has to be involved. Actually, if we are not focusing on faulty states, $(D)$ $(0.699)$ is the best singleton explanation. It is clearly not an adequate explanation for the evidence.

For a domain in which target variables are interdependent, multivariate explanations are often more natural for explaining given evidence. However, existing methods often produce hypotheses that are *overspecified*. MAP finds a configuration of a set of target variables that maximize the joint posterior probability given partial evidence on the other variables. For the running example, if we set $A, B, C$ and $D$ as the target variables, MAP will find $(A \wedge \neg B \wedge \neg C \wedge D)$ as the best explanation. However, given that $B$ and $C$ are faulty, $A$ and $D$ are somewhat redundant for explaining the evidence. MPE finds an explanation with even more variables. Several other approaches use the dependence relations encoded in Bayesian networks to prune independent variables [10, 11]. They will find the same explanation as MAP because all of the target variables are dependent on the evidence. Yet several other methods measure the quality of an explanation using the likelihood of the evidence [1]. Unfortunately they will overfit and choose $(\neg A \wedge \neg B \wedge \neg C \wedge \neg D)$ as the explanation, because the likelihood of the evidence given that all the target variables fail is almost $1.0$.

There have been efforts trying to generate more appropriate explanations. Henrion and Druzdzel [6] assume that a system has a set of pre-defined scenarios as potential explanations and find the scenario with the highest posterior probability. Flores et al. [4] propose to grow an explanation tree incrementally by branching the most informative variable at each step while maintaining the probability of each explanation above certain threshold. Nielsen et al. [8] use a different measure called causal information flow to grow the explanation trees. Because the explanations in the trees have to branch on the same variable(s), they may still contain redundant variables. Finding more concise hypotheses also have been studied in model-based diagnosis [2]. The approach focus on truth-based systems and cannot be easily generalized to deal with Bayesian networks.

---

[1] We use a variable and its negation to stand for its "ok" and "defective" states respectively

## 4 Most Relevant Explanation

There are two most essential properties for a good explanation. First, the explanation should be *precise*, meaning it should explain the presence of the evidence well. Second, the explanation should be *concise* and only contain the most relevant variables. The above discussions show that existing approaches for explaining evidence in Bayesian networks often generate explanations that are either underspecified (imprecise) or overspecified (inconcise).

To address the limitations, Yuan and Lu [12] propose a method called *Most Relevant Explanation* (MRE) to automatically identify the most relevant target variables for given evidence in Bayesian networks. First, *explanation* in Bayesian networks is formally defined as follows.

**Definition 1.** *Given a set of target variables* $\mathbf{X}$ *in a Bayesian network and evidence* $\mathbf{e}$ *on the remaining variables, an* explanation *for the evidence is a partial instantiation* $\mathbf{x_{1:k}}$ *of* $\mathbf{X}$*, i.e.,* $\mathbf{X_{1:k}} \subseteq \mathbf{X}$ *and* $\mathbf{X_{1:k}} \neq \emptyset$.

MRE is then defined as follows [12].

**Definition 2.** *Let* $\mathbf{X}$ *be a set of target variables, and* $\mathbf{e}$ *be the evidence on the remaining variables in a Bayesian network.* Most Relevant Explanation *is the problem of finding an explanation* $\mathbf{x_{1:k}}$ *that has the maximum* Generalized Bayes Factor *score* $GBF(\mathbf{x_{1:k}}; \mathbf{e})$, *i.e.,*

$$MRE(\mathbf{X}, \mathbf{e}) \equiv \arg\max_{\mathbf{x_{1:k}}, \mathbf{X_{1:k}} \subseteq \mathbf{X}, \mathbf{X_{1:k}} \neq \emptyset} GBF(\mathbf{x_{1:k}}; \mathbf{e}), \quad (1)$$

*where* $GBF$ *is defined as*

$$GBF(\mathbf{x_{1:k1}}; \mathbf{e}) \equiv \frac{P(\mathbf{e}|\mathbf{x_{1:k1}})}{P(\mathbf{e}|\overline{\mathbf{x_{1:k1}}})}. \quad (2)$$

Therefore, MRE traverses the trans-dimensional space containing all the partial assignments of $\mathbf{X}$ and finds an assignment that maximizes the $GBF$ score. Potentially, MRE can use any measure that provides a common ground for comparing the partial instantiations of the target variables. $GBF$ is chosen because it is shown to provide a plausible measure for representing the degree of evidential support in recent studies in Bayesian confirmation theory [3].

MRE was shown to be able to generate precise and concise explanations for the running example [12]. The best explanation according to MRE is:

$$GBF(\neg B, \neg C; \mathbf{e}) = 42.62. \quad (3)$$

For simplicity we often omit $\mathbf{e}$ and write $GBF(\neg B, \neg C)$. $(\neg B, \neg C)$ is a better explanation than both $(\neg A)$ (39.44) and $(\neg B, \neg D)$ (35.88), because its prior and posterior probabilities are both relatively high; The posterior probabilities of the explanations are 0.394, 0.391, and 0.266 respectively. Therefore, MRE seems able to automatically identify the most relevant target variables and states as the explanations for given evidence.

## 5 A Theoretical Study

### 5.1 Theoretical properties of MRE

We now discuss several theoretical properties of MRE. Since MRE relies heavily on the $GBF$ measure in generating its explanations, it is not surprising that these properties are mostly originated from $GBF$. The proofs of these properties can be found in the appendix.

First, we note that GBF can be expressed in a different way using the *belief update ratio*.

**Definition 3.** *The* belief update ratio *of* $\mathbf{x_{1:k1}}$ *given* $\mathbf{e}$, $r(\mathbf{x_{1:k1}}; \mathbf{e})$, *is defined as*

$$r(\mathbf{x_{1:k}}; \mathbf{e}) \equiv \frac{P(\mathbf{x_{1:k}}|\mathbf{e})}{P(\mathbf{x_{1:k}})}. \quad (4)$$

GBF can then be expressed as the ratio between the belief update ratios of $\mathbf{x_{1:k1}}$ and alternative explanations $\overline{\mathbf{x_{1:k1}}}$ given $\mathbf{e}$, i.e.,

$$GBF(\mathbf{x_{1:k1}}; \mathbf{e}) = \frac{r(\mathbf{x_{1:k1}}; \mathbf{e})}{r(\overline{\mathbf{x_{1:k1}}}; \mathbf{e})}. \quad (5)$$

The most important property of MRE is that it is able to weigh the relative importance of multiple variables and only include the most relevant variables in explaining the given evidence. The degree of relevance is evaluated using a measure called *conditional Bayes factor* (CBF) implicitly encoded in the GBF measure and defined as follows.

**Definition 4.** *The* conditional Bayes factor *of hypothesis* $\mathbf{y_{1:m}}$ *for given evidence* $\mathbf{e}$ *conditional on* $\mathbf{x_{1:k}}$ *is defined as*

$$CBF(\mathbf{y_{1:m}}; \mathbf{e}|\mathbf{x_{1:k}}) \equiv \frac{P(\mathbf{e}|\mathbf{y_{1:m}}, \mathbf{x_{1:k}})}{P(\mathbf{e}|\overline{\mathbf{y_{1:m}}}, \mathbf{x_{1:k}})}. \quad (6)$$

Then, we have the following theorem.

**Theorem 1.** *Let the conditional Bayes factor of* $\mathbf{y_{1:m}}$ *given* $\mathbf{x_{1:k}}$ *be less than or equal to inverse of the belief update ratio of the alternative explanations* $\overline{\mathbf{x_{1:k}}}$, *i.e.,*

$$CBF(\mathbf{y_{1:m}}; \mathbf{e}|\mathbf{x_{1:k}}) \leq \frac{1}{r(\overline{\mathbf{x_{1:k}}}; \mathbf{e})}, \quad (7)$$

*the following holds*

$$GBF(\mathbf{x_{1:k}} \cup \mathbf{y_{1:m}}; \mathbf{e}) \leq GBF(\mathbf{x_{1:k}}; \mathbf{e}). \quad (8)$$

Therefore, $CBF(\mathbf{y_{1:m}}, \mathbf{e}|\mathbf{x_{1:k}})$ provides a soft measure on the relevance of a new set of variable states with regard to an existing explanation and can be used to decide whether or not to include them in an existing explanation. $GBF$ also encodes a decision boundary, the inverse belief update ratio of alternative explanations $\overline{\mathbf{x_{1:k}}}$ given $\mathbf{e}$, which provides a threshold on how important the remaining variables should be in order to be included in the current explanation. If $CBF(\mathbf{y_{1:m}}; \mathbf{e}|\mathbf{x_{1:k}})$ is greater than or equal to $\frac{1}{r(\overline{\mathbf{x_{1:k}}}; \mathbf{e})}$, $\mathbf{y_{1:m}}$ is regarded as relevant and will be included. Otherwise, $\mathbf{y_{1:m}}$ will be excluded from the explanation.

Theorem 1 has several intuitive and desirable corollaries. First, the following corollary shows that, for any explanation $x_{1:k}$ with belief update ratio greater than or equal to 1.0, adding any independent variable to the explanation will decrease its GBF score [12].

**Corollary 1.** *Let $\mathbf{x_{1:k}}$ be an explanation with $r(\mathbf{x_{1:k}}; \mathbf{e}) \geq 1.0$, and $y$ be a state of variable $Y$ independent from variables in $\mathbf{x_{1:k}}$ and $\mathbf{e}$. Then*

$$GBF(\mathbf{x_{1:k}} \cup \{y\}; \mathbf{e}) \leq GBF(\mathbf{x_{1:k}}; \mathbf{e}). \quad (9)$$

Therefore, adding an irrelevant variable dilutes the explanative power of an existing explanation. MRE is able to automatically prune such variables. This is clearly a desirable property.

Note that we focus on the explanations with belief update ratio greater than or equal to 1.0. We believe that an explanation whose probability decreases given the evidence is unlikely to be a good explanation for the evidence.

Corollary 1 requires the additional variable $Y$ to be independent from both $\mathbf{X_{1:k}}$ and $\mathbf{E}$. The assumption is rather strong. The following corollary relaxes it to be that $Y$ is conditionally independent from $\mathbf{E}$ given $\mathbf{X_{1:k}}$ and shows the same result still holds.

**Corollary 2.** *Let $\mathbf{x_{1:k}}$ be an explanation with $r(\mathbf{x_{1:k}}; \mathbf{e}) \geq 1.0$, and $y$ be a state of a variable $Y$ conditionally independent from variables in $\mathbf{e}$ given $\mathbf{x_{1:k}}$. Then*

$$GBF(\mathbf{x_{1:k}} \cup \{y\}; \mathbf{e}) \leq GBF(\mathbf{x_{1:k}}; \mathbf{e}). \quad (10)$$

Corollary 2 is a more general result than corollary 1 and captures the intuition that conditionally independent variables add no additional information to an explanation in explaining given evidence, even though the variable may be marginally dependent on the evidence. Also note that these properties are all relative to an existing explanation. It is possible that a variable is independent from the evidence given one explanation, but becomes dependent on the evidence given another explanation. In other words, GBF score is not monotonic. Looking at variables one by one does not guarantee to find the optimal solution.

The above results can be further relaxed to accommodate cases where the posterior probability of $y$ given $\mathbf{e}$ is smaller than its prior, i.e.,

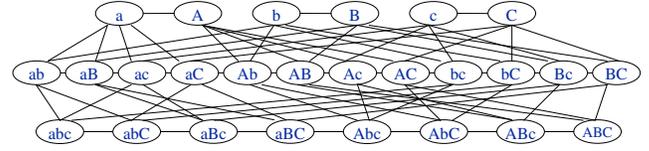

Figure 2: Solution space of Most Relevant Explanation

**Corollary 3.** *Let $\mathbf{x_{1:k}}$ be an explanation with $r(\mathbf{x_{1:k}}; \mathbf{e}) \geq 1.0$, and $y$ be a state of a variable $Y$ such that $P(y|\mathbf{x_{1:k}}, \mathbf{e}) \leq P(y|\mathbf{x_{1:k}})$. Then*

$$GBF(\mathbf{x_{1:k}} \cup \{y\}; \mathbf{e}) \leq GBF(\mathbf{x_{1:k}}; \mathbf{e}). \quad (11)$$

This is again an intuitive result; a variable state whose posterior probability decreases for given evidence should not be part of an explanation for the evidence.

The above theoretical results can be verified using the running example. For example,

$$\begin{aligned} & GBF(\neg B, \neg C) \\ > \; & GBF(\neg B, \neg C, A) \;\&\; GBF(\neg B, \neg C, D) \\ > \; & GBF(\neg B, \neg C, A, D) \,. \end{aligned}$$

The results suggest that GBF has the intrinsic capability to penalize higher-dimensional explanations and prune less relevant variables.

### 5.2 Explaining away

One unique property of Bayesian networks is that they can model the so called *explaining away* phenomenon using the $V$ structure, i.e., a single variable with two or more parents. This structure intuitively captures the situation where an effect has multiple causes. Observing the presence of the effect and one of the causes reduces the likelihood of the presence of the other causes. It is desirable to capture this phenomenon when generating explanations.

MRE seems able to capture the explaining away effect using CBF. CBF provides a measure on how relevant a new variable is to an existing explanation. In an explaining-away situation, if one of the causes is already present in the current explanation, other causes typically do not receive high CBF scores. Again for the running example, $(\neg B, \neg C)$ and $(\neg A)$ are both good explanations for the evidence by themselves. The $CBF$ of $\neg A$ given only $\mathbf{e}$ (the effect) is equal to its GBF (39.44), which is rather high. However, when $(\neg B, \neg C)$ (one of the causes) is also observed, $CBF(\neg A; \mathbf{e}|\neg B, \neg C)$ becomes rather low and is only equal to 1.03. Clearly, CBF is able to capture the explaining away phenomenon in this example.

## 5.3 Dominance relations

MRE has a solution space with an interesting lattice structure similar to the graph in Figure 2 for three binary target variables. The graph contains all the partial assignments of the target variables. Two explanations are linked together if they only have a local difference, meaning they either have the same set of variables with one variable in different states, or one explanation has one fewer variable than the other explanation with all the other variables being in the same states.

There are two dominance relations among these potential solutions that are implied by Figure 2. The first concept is *strong dominance*.

**Definition 5.** *An explanation* $\mathbf{x}_{1:k}$ strongly dominates *another explanation* $\mathbf{y}_{1:m}$ *if and only if* $\mathbf{x}_{1:k} \subset \mathbf{y}_{1:m}$ *and* $GBF(\mathbf{x}_{1:k}) \geq GBF(\mathbf{y}_{1:m})$.

If $\mathbf{x}_{1:k}$ strongly dominates $\mathbf{y}_{1:m}$, $\mathbf{x}_{1:k}$ is clearly a better explanation than $\mathbf{y}_{1:m}$, because it not only has a no-worse explanative score but also is more concise. We only need to consider $\mathbf{x}_{1:k}$ when finding multiple top MRE explanations. The second concept is *weak dominance*.

**Definition 6.** *An explanation* $\mathbf{x}_{1:k}$ weakly dominates *another explanation* $\mathbf{y}_{1:m}$ *if and only if* $\mathbf{x}_{1:k} \supset \mathbf{y}_{1:m}$ *and* $GBF(\mathbf{x}_{1:k}) > GBF(\mathbf{y}_{1:m})$.

In this case, $\mathbf{x}_{1:k}$ has a strictly larger $GBF$ score than $\mathbf{y}_{1:m}$, but the latter is more concise. It is possible that we can include them both and let the decision makers to decide whether they prefer higher score or conciseness. However, we believe that we only need to include $\mathbf{x}_{1:k}$, because its higher GBF score indicates that the extra variable states are relevant to explain given evidence and should be included in the explanation.

Based on the two kinds of dominance relations, we define the concept *minimal*.

**Definition 7.** *An explanation is* minimal *if it is neither strongly nor weakly dominated by any other explanation.*

In case we want to find multiple top explanations, we only need to consider the minimal explanations, because they are the most representative ones.

## 6 K-MRE Algorithm

In many decision problems, outputting the single top solution may not be the best practice. Decision makers typically would like multiple competing options to choose from. This is especially important when there are multiple solutions that are almost equally good. For the circuit example, all three basic explanations will lead to the same observation. However, we can only recover one explanation if we are satisfied with one top solution. It is better to output all the top solutions rather than selecting any one of the solutions.

| GBF(¬ B, ¬ C) = 42.62 | GBF(¬ A) = 39.44 |
| GBF(A, ¬ B, ¬ C) = 42.15 | GBF(¬ A, B) = 36.98 |
| GBF(¬ B, ¬ C, D) = 39.93 | GBF(¬ A, C) = 35.99 |
| GBF(A, ¬ B, ¬ C, D) = 39.56 | **GBF(¬ B, ¬ D) = 35.88** |

Table 1: The top solutions ranked by GBF. The solutions in boldface are the top minimal solutions.

The dominance relations defined in the last section allow us to develop a K-MRE algorithm to find a set of top solutions that are more representative. Let us look at the running example again to illustrate the idea. The explanations in Table 1 have the highest GBF scores. If we simply select top three explanations solely based on GBF, we will obtain these rather similar explanations: $(\neg B, \neg C)$, $(A, \neg B, \neg C)$, and $(\neg B, \neg C, D)$, which are rather similar. Since $(A, \neg B, \neg C)$, $(\neg B, \neg C, D)$, and $(A, \neg B, \neg C, D)$ are strongly dominated by $(\neg B, \neg C)$, we should only consider $(\neg B, \neg C)$ out of those four explanations. Similarly, $(\neg A, B)$ and $(\neg A, C)$ are strongly dominated by $(\neg A)$. These dominated explanations should be excluded from the top solution set. In the end, we get the set of top explanations shown in boldface in Table 1, which is clearly more diverse and representative than the original set. MAP and MPE clearly do not have this nice property.

Therefore, our proposed K-MRE algorithm works as follows. Whenever we generate a new explanation, we check its score against the best solution pool. If it is lower than the worst score in the pool, reject the new explanation. If there are fewer than $K$ best solutions or if the score of the new explanation is higher than the worst score in the pool, we consider adding the new explanation to the top pool. We first check whether the new solution is strongly or weakly dominated by any of the top explanations. If so, reject the new explanation. Otherwise, we add the new explanation to the top pool. However, we then need to check whether there are existing top explanations that are dominated by the newly added explanation. If yes, these existing explanations should be excluded. Otherwise we delete the top explanation with the least score.

## 7 Empirical Results

### 7.1 Experimental design

We tested the K-MRE algorithm on a set of benchmark models, including Alarm, Circuit, Hepar, Munin, and SmallHepar. We chose these several models because we have the diagnostic versions of these networks, whose variables have been annotated into three categories: *target*, *observation*, and *auxiliary*. For generating the test cases, we used the networks as generative models and sampled *with-*

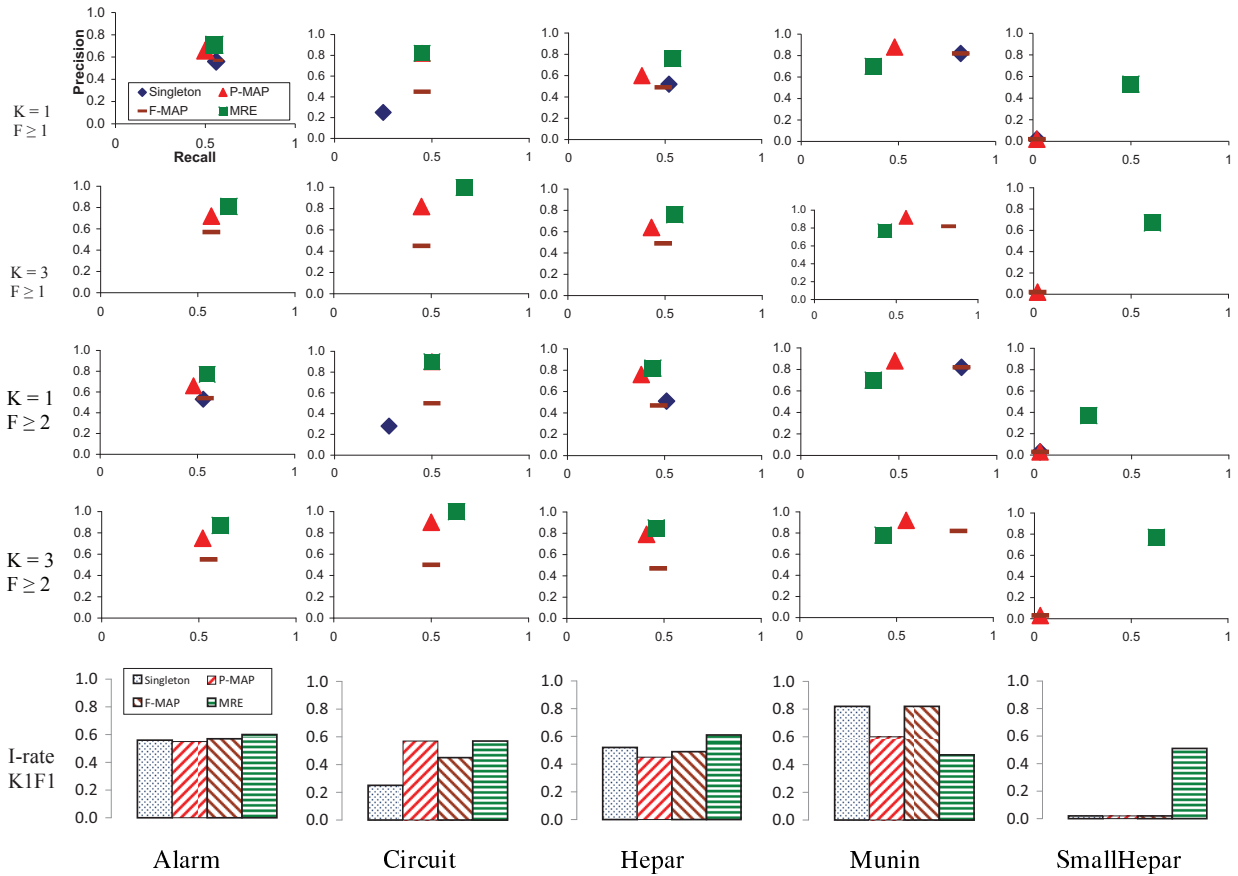

Figure 3: Precision vs recall plots of the results by four algorithms, Marginal, P-MAP, F-MAP, and MRE, on a set of benchmark diagnostic Bayesian networks. "K" shows the number of top solutions generated. "F" shows the least number of faulty target variables in test cases. "F_Score" shows the F-Scores of the results of the algorithms. Marginal algorithm did not appear in rows "K3F1" and "K3F2" because it has only one solution.

*out replacement* from their prior probability distributions. We only kept those test cases with at least one abnormal observation and used the abnormal observations as evidence. Since Circuit and SmallHepar have 4 and 3 target variables respectively, we collected as many test cases as possible. Munin also has 4 target variables but each with many more states. Hepar and Alarm have 9 and 12 target variables respectively. We collected 50 test cases for the last three networks. We also extracted from them the test cases which contain at least two faulty target variables for separate experiments on multiple-fault test cases.

Our experiments compared MRE with MAP given their similarities. We tested two versions of the MAP algorithm, one focusing on all the target variables (F-MAP) and the other only on the target variables selected by MRE (P-MAP). In addition, we compared with the *Marginal* algorithm, which neglects the interdependence among the target variables and uses the marginal posterior probabilities to determine the most likely states of the target variables. We plot the accuracy statistics, including *precision* (the percentage of faulty states correctly identified among all faulty explanation variables) and *recall* (the percentage of faulty states correctly identified among all faulty variables in test cases) of these algorithms in Figure 3. We also include sample results on *F-Score*, which is defined as

$$\text{F-Score} = \frac{2 \times (precision \times recall)}{(precision + recall)}. \quad (12)$$

### 7.2 Results and analysis

We make the following observations from these results. First, MRE is able to achieve higher precision and/or recall rates in identifying the faulty target variables than the other algorithms on all the networks except Munin. An outstanding example is the SmallHepar network. Marginal, F-MAP and P-MAP all failed badly on this model in identifying the faulty variables, while MRE was able to achieve reasonable performance. It is clearly desirable given that one major goal of diagnosis or explanation is to identify problems, e.g. faulty states. We investigated the results of Munin network further and found that all target variables of these test cases are in faulty states. Marginal and F-MAP have exactly the same statistics, which suggests that the target variables may have weak correlations with each other. This puts MRE in disadvantage because MRE takes into account such weak correlations and generate concise explanations with fewer target variables. On average, the explanations of MRE identifies 4.3 variables out of 12 target variables for Alarm, 1.7/4 for Circuit, 4/9 for Hepar, 2.5/4 for Munin, and 2.3/3 for SmallHepar. For networks with strong correlations among the target variables, e.g. Circuit and Hepar, MRE has much higher precision/recall rates. The sample F-score results in the case of "K1F1" further confirmed the observation.

Second, by comparing rows "K1F1" vs. "K3F1" and "K1F2" vs. "K3F2," we found that using multiple top solutions helps MRE significantly in improving the precision/recall rates than the other algorithms. With multiple solutions, we kept the results with the maximum precision rates. The results seem to support our claim that K-MRE was able to generate solutions that are more representative. It is somewhat surprising that the precision/recall rates of F-MAP were not improved at all on the networks, but those of P-MAP were improved. Our hypothesis is that, since the explanations by F-MAP are more grained because more variables are involved, its top explanations tend to agree with each other on the faulty variables and differ mostly in the less important non-faulty variables. Generating multiple top solutions could not really help F-MAP much in improving its accuracy statistics.

Third, although P-MAP gets the target variables identified by MRE as input, it still failed badly on the SmallHepar network in identifying faulty states of the target variables. It did not show any significant advantage over F-MAP on other networks either. The results suggest that relying on posterior probabilities may not work well in certain diagnostic systems.

Fourth, although multiple-fault cases are believed to be more difficult because of their low likelihood, the algorithms in our experiments seem able to maintain the same level of accuracy rates in face of multiple-fault test cases (rows "K1F2" and "K3F2"). We hope to apply the proposed methods to real-world systems and test cases to gain more insights.

Last but not least, the Marginal algorithm is efficient and sometimes can achieve similar accuracy rates with other methods. However, since it does not take into account the dependence among the target variables, its results can be arbitrarily bad if the dependence are strong. It is evident on the Circuit network for which the accuracy rates of Marginal algorithm are much lower than other methods. The results suggest that we have to be cautious about the use of the Marginal algorithm in certain systems.

## 8 Concluding Remarks

In this paper, we discuss several theoretical properties of Most Relevant Explanation (MRE) and develop an algorithm for finding multiple top MRE solutions. Our study shows that MRE relies on an implicit soft relevance measure in automatically identifying the most relevant target variables and pruning less relevant variables from an explanation. The soft measure also enables MRE to capture the intuitive phenomenon of explaining away encoded in Bayesian networks. Furthermore, we define two dominance relations among the explanations that are implied by the structure of the solution space of MRE. These relations allow us to design and develop a K-MRE algorithm for finding top MRE solutions that are much more representative.

Our empirical results agree quite well with the theoretical understanding of MRE. The results show that MRE is effective in identifying the most relevant target variables, especially the true faulty target variables. Furthermore, K-MRE seems able to generate more representative top explanations than K-MAP methods. We believe that MRE is especially suitable for systems in which target variables are strong correlated with each other and can generate more precise and concise explanations for these systems.

This research has many future works. It is desirable to understand the theoretical complexity of MRE. It has a solution space even larger than MAP and is believed to be at least as hard. Currently we rely on an exhaustive search algorithm for solving MRE and K-MRE. More efficient methods for solving MRE need be developed to make it applicable to large real-world problems.

**Acknowledgement**  This research was supported by the National Science Foundation grant IIS-0842480. All experimental data have been obtained using SMILE, a Bayesian inference engine developed at the Decision Systems Laboratory at University of Pittsburgh and available at http://genie.sis.pitt.edu.

## References

[1] U. Chajewska and J. Y. Halpern. Defining explanation in probabilistic systems. In *Proceedings of the Thirteenth Annual Conference on Uncertainty in Artificial Intelligence (UAI–97)*, pages 62–71, San Francisco, CA, 1997. Morgan Kaufmann Publishers.

[2] J. de Kleer and B. C. Willams. Diagnosis with behavioral modes. In *Proceedings of IJCAI-89*, pages 104–109, 1989.

[3] B. Fitelson. Likelihoodism, Bayesianism, and relational confirmation. *Synthese*, 156(3), June 2007.

[4] J. Flores, J. A. Gamez, and S. Moral. Abductive inference in Bayesian networks: finding a partition of the explanation space. In *Eighth European Conference on Symbolic and*

*Quantitative Approaches to Reasoning with Uncertainty, ECSQARU'05*, pages 63–75. Springer Verlag, 2005.

[5] D. E. Heckerman, E. J. Horvitz, , and B. N. Nathwani. Toward normative expert systems: Part I the pathfinder project. *Methods of Information in Medicine*, 31:90–105, 1992.

[6] M. Henrion and M. J. Druzdzel. Qualitative propagation and scenario-based schemes for explaining probabilistic reasoning. In P. Bonissone, M. Henrion, L. Kanal, and J. Lemmer, editors, *Uncertainty in Artificial Intelligence 6*, pages 17–32. Elsevier Science Publishing Company, Inc., New York, N. Y., 1991.

[7] J. Kalagnanam and M. Henrion. A comparison of decision analysis and expert rules for sequential diagnosis. In *Proceedings of the 4th Annual Conference on Uncertainty in Artificial Intelligence (UAI-88)*, pages 253–270, New York, NY, 1988. Elsevier Science.

[8] U. Nielsen, J.-P. Pellet, and A. Elisseeff. Explanation trees for causal Bayesian networks. In *Proceedings of the 24th Annual Conference on Uncertainty in Artificial Intelligence (UAI-08)*, 2008.

[9] D. Poole and G. M. Provan. What is the most likely diagnosis? In P. Bonissone, M. Henrion, L. Kanal, and J. Lemmer, editors, *Uncertainty in Artificial Intelligence 6*, pages 89–105. Elsevier Science Publishing Company, Inc., New York, N. Y., 1991.

[10] S. E. Shimony. The role of relevance in explanation I: Irrelevance as statistical independence. *International Journal of Approximate Reasoning*, 8(4):281–324, June 1993.

[11] L. van der Gaag and M. Wessels. *Efficient multiple-disorder diagnosis by strategic focusing*, pages 187–204. UCL Press, London, 1995.

[12] C. Yuan and T.-C. Lu. A general framework for generating multivariate explanations in Bayesian networks. In *Proceedings of the Twenty-Third National Conference on Artificial Intelligence (AAAI-08)*, 2008.

**Appendix**

**Proof of Theorem 1:**

$$GBF(\mathbf{x_{1:k}} \cup \mathbf{y_{1:m}}; \mathbf{e})$$
$$= \frac{P(\mathbf{x_{1:k}} \cup \mathbf{y_{1:m}}|\mathbf{e})(1 - P(\mathbf{x_{1:k}} \cup \mathbf{y_{1:m}}))}{P(\mathbf{x_{1:k}} \cup \mathbf{y_{1:m}})(1 - P(\mathbf{x_{1:k}} \cup \mathbf{y_{1:m}}|\mathbf{e}))}$$
$$= \frac{P(\mathbf{x_{1:k}}|\mathbf{e})P(\mathbf{y_{1:m}}|\mathbf{x_{1:k}}, \mathbf{e})(1 - P(\mathbf{y_{1:m}}|\mathbf{x_{1:k}})P(\mathbf{x_{1:k}}))}{P(\mathbf{x_{1:k}})P(\mathbf{y_{1:m}}|\mathbf{x_{1:k}})(1 - P(\mathbf{y_{1:m}}|\mathbf{x_{1:k}}, \mathbf{e})P(\mathbf{x_{1:k}}|\mathbf{e}))}$$
$$= \frac{P(\mathbf{x_{1:k}}|\mathbf{e})}{P(\mathbf{x_{1:k}})} \frac{1 - P(\mathbf{x_{1:k}}) + \frac{1}{P(\mathbf{y_{1:m}}|\mathbf{x_{1:k}})} - 1}{1 - P(\mathbf{x_{1:k}}|\mathbf{e}) + \frac{1}{P(\mathbf{y_{1:m}}|\mathbf{x_{1:k}},\mathbf{e})} - 1}$$

The above equation is less than or equal to $GBF(\mathbf{x_{1:k}}; \mathbf{e})$ when

$$\frac{\frac{1}{P(\mathbf{y_{1:m}}|\mathbf{x_{1:k}})} - 1}{\frac{1}{P(\mathbf{y_{1:m}}|\mathbf{x_{1:k}},\mathbf{e})} - 1} \leq \frac{1 - P(\mathbf{x_{1:k}})}{1 - P(\mathbf{x_{1:k}}|\mathbf{e})}$$
$$\Leftrightarrow \frac{P(\mathbf{y_{1:m}}|\mathbf{x_{1:k}}, \mathbf{e})(1 - P(\mathbf{y_{1:m}}|\mathbf{x_{1:k}}))}{P(\mathbf{y_{1:m}}|\mathbf{x_{1:k}})(1 - P(\mathbf{y_{1:m}}|\mathbf{x_{1:k}}, \mathbf{e}))} \leq \frac{P(\overline{\mathbf{x_{1:k}}})}{P(\overline{\mathbf{x_{1:k}}}|\mathbf{e})}$$
$$\Leftrightarrow CBF(\mathbf{y_{1:m}}; \mathbf{e}|\mathbf{x_{1:k}}) \leq \frac{1}{r(\overline{\mathbf{x_{1:k}}}; \mathbf{e})}.$$

**Proof of Corollary 1:** The corollary follows immediately from Theorem 1. We can also prove it in the following way.

$$GBF(\mathbf{x_{1:k}} \cup \{y\}; \mathbf{e})$$
$$= \frac{P(\mathbf{x_{1:k}} \cup \{y\}|\mathbf{e})(1 - P(\mathbf{x_{1:k}} \cup \{y\}))}{P(\mathbf{x_{1:k}} \cup \{y\})(1 - P(\mathbf{x_{1:k}} \cup \{y\}|\mathbf{e}))}$$
$$= \frac{P(\mathbf{x_{1:k}}|\mathbf{e})P(y)(1 - P(y)P(\mathbf{x_{1:k}}))}{P(\mathbf{x_{1:k}})P(y)(1 - P(y)P(\mathbf{x_{1:k}}|\mathbf{e}))}$$
$$= \frac{P(\mathbf{x_{1:k}}|\mathbf{e})(1 - P(y)P(\mathbf{x_{1:k}}))}{P(\mathbf{x_{1:k}})(1 - P(y)P(\mathbf{x_{1:k}}|\mathbf{e}))}.$$

Because $P(\mathbf{x_{1:k}}|\mathbf{e}) \geq P(\mathbf{x_{1:k}})$, we have the following:

$$GBF(\mathbf{x_{1:k}} \cup \{y\}; \mathbf{e})$$
$$= \frac{P(\mathbf{x_{1:k}}|\mathbf{e})(1 - P(y)P(\mathbf{x_{1:k}}))}{P(\mathbf{x_{1:k}})(1 - P(y)P(\mathbf{x_{1:k}}|\mathbf{e}))}$$
$$= \frac{P(\mathbf{x_{1:k}}|\mathbf{e})(1 - P(\mathbf{x_{1:k}}) + (1 - p(y))P(\mathbf{x_{1:k}}))}{P(\mathbf{x_{1:k}})(1 - P(\mathbf{x_{1:k}}|\mathbf{e}) + (1 - p(y))P(\mathbf{x_{1:k}}|\mathbf{e}))}$$
$$\leq \frac{P(\mathbf{x_{1:k}}|\mathbf{e})(1 - P(\mathbf{x_{1:k}}) + (1 - p(y))P(\mathbf{x_{1:k}}))}{P(\mathbf{x_{1:k}})(1 - P(\mathbf{x_{1:k}}|\mathbf{e}) + (1 - p(y))P(\mathbf{x_{1:k}}))}$$
$$\leq \frac{P(\mathbf{x_{1:k}}|\mathbf{e})(1 - P(\mathbf{x_{1:k}}))}{P(\mathbf{x_{1:k}})(1 - P(\mathbf{x_{1:k}}|\mathbf{e}))}$$
$$= GBF(\mathbf{x_{1:k}}; \mathbf{e}).$$

**Proof of Corollary 2:** This corollary can be proved in a similar way as Corollary 1.

$$GBF(\mathbf{x_{1:k}} \cup \{y\}; \mathbf{e})$$
$$= \frac{P(\mathbf{x_{1:k}} \cup \{y\}|\mathbf{e})(1 - P(\mathbf{x_{1:k}} \cup \{y\}))}{P(\mathbf{x_{1:k}} \cup \{y\})(1 - P(\mathbf{x_{1:k}} \cup \{y\}|\mathbf{e})}$$
$$= \frac{P(\mathbf{x_{1:k}}|\mathbf{e})P(y|\mathbf{x_{1:k}}, \mathbf{e})(1 - P(y|\mathbf{x_{1:k}}, \mathbf{e})P(\mathbf{x_{1:k}}))}{P(\mathbf{x_{1:k}})P(y|\mathbf{x_{1:k}})(1 - P(y|\mathbf{x_{1:k}})P(\mathbf{x_{1:k}}|\mathbf{e}))}$$
$$= \frac{P(\mathbf{x_{1:k}}|\mathbf{e})P(y|\mathbf{x_{1:k}})(1 - P(y|\mathbf{x_{1:k}})P(\mathbf{x_{1:k}}))}{P(\mathbf{x_{1:k}})P(y|\mathbf{x_{1:k}})(1 - P(y|\mathbf{x_{1:k}})P(\mathbf{x_{1:k}}|\mathbf{e}))}$$
$$= \frac{P(\mathbf{x_{1:k}}|\mathbf{e})(1 - P(y|\mathbf{x_{1:k}})P(\mathbf{x_{1:k}}))}{P(\mathbf{x_{1:k}})(1 - P(y|\mathbf{x_{1:k}})P(\mathbf{x_{1:k}}|\mathbf{e})}.$$

Because $P(\mathbf{x_{1:k}}|\mathbf{e}) \geq P(\mathbf{x_{1:k}})$, we have

$$GBF(\mathbf{x_{1:k}} \cup \{y\}; \mathbf{e})$$
$$= \frac{P(\mathbf{x_{1:k}}|\mathbf{e})(1 - P(y|\mathbf{x_{1:k}})P(\mathbf{x_{1:k}}))}{P(\mathbf{x_{1:k}})(1 - P(y|\mathbf{x_{1:k}})P(\mathbf{x_{1:k}}|\mathbf{e})}$$
$$= \frac{P(\mathbf{x_{1:k}}|\mathbf{e})(1 - P(\mathbf{x_{1:k}}) + (1 - p(y|\mathbf{x_{1:k}}))P(\mathbf{x_{1:k}}))}{P(\mathbf{x_{1:k}})(1 - P(\mathbf{x_{1:k}}|\mathbf{e}) + (1 - p(y|\mathbf{x_{1:k}}))P(\mathbf{x_{1:k}}|\mathbf{e}))}$$
$$\leq \frac{P(\mathbf{x_{1:k}}|\mathbf{e})(1 - P(\mathbf{x_{1:k}}) + (1 - p(y|\mathbf{x_{1:k}}))P(\mathbf{x_{1:k}}))}{P(\mathbf{x_{1:k}})(1 - P(\mathbf{x_{1:k}}|\mathbf{e}) + (1 - p(y|\mathbf{x_{1:k}}))P(\mathbf{x_{1:k}}))}$$
$$\leq \frac{P(\mathbf{x_{1:k}}|\mathbf{e})(1 - P(\mathbf{x_{1:k}}))}{P(\mathbf{x_{1:k}})(1 - P(\mathbf{x_{1:k}}|\mathbf{e}))}$$
$$= GBF(\mathbf{x_{1:k}}; \mathbf{e}).$$

**Proof of Corollary 3:**

This corollary follows immediately from Theorem 1.